\documentclass[useAMS,usenatbib,referee]{biom}
\def\beqnn{\begin{eqnarray*}}
\def\eeqnn{\end{eqnarray*}}
\def\beqlb{\begin{eqnarray}}
\def\eeqlb{\end{eqnarray}}

\usepackage{graphics}
\usepackage{graphicx}
\usepackage{epsf}
\usepackage{epstopdf}
\usepackage{epsfig}
\usepackage{amsmath}
\usepackage{amsfonts}
\usepackage{hyperref}

\usepackage[figuresright]{rotating}
\title{Equivalence of Kernel Machine Regression and Kernel Distance Covariance for Multidimensional Trait Association Studies$^{\ast}$}
\author{Wen-Yu Hua$^{1}$ and Debashis Ghosh$^{2}$ \\
\small{$^1$Department of Statistics, The Pennsylvania State University;} \\
\small{$^{2}$Departments of Statistics and Public Health Sciences, The Pennsylvania State University} \\
}

\begin{document}
%
%
%

\label{firstpage}
\begin{abstract}
{

Associating genetic markers with a multidimensional phenotype is an important yet challenging problem. In this work, we establish the equivalence between two popular methods: kernel-machine regression (KMR), and kernel distance covariance (KDC). KMR is a semiparametric regression frameworks that models the covariate effects parametrically, while the genetic markers are considered non-parametrically.  KDC represents a class of methods that includes distance covariance (DC) and Hilbert-Schmidt Independence Criterion (HSIC), which are nonparametric tests of independence. We show the equivalence between the score test of KMR and the KDC statistic under certain conditions.  This result leads to a novel generalization of the KDC test that incorporates the covariates.
Our contributions are three-fold: (1) establishing the equivalence between KMR and KDC; (2) showing that the principles of kernel machine regression can be applied to the interpretation of KDC; (3) the development of a broader class of KDC statistics, that the members are the quantities of different kernels. We demonstrate the proposals using simulation studies.   Data from the Alzheimer's Disease Neuroimaging Initiative (ADNI) is used to explore the association between the genetic variants on gene \emph{FLJ16124} and phenotypes represented in 3D structural brain MR images adjusting for age and gender. The results suggest that SNPs of \emph{FLJ16124} exhibit strong pairwise interaction effects that are correlated to the changes of brain region volumes.


\hrulefill\\
\footnotesize{$^{\ast}$ Data used in preparation of this article were obtained from the Alzheimer's Disease Neuroimaging Initiative (ADNI) database (adni.loni.usc.edu). As such, the investigators within the ADNI contributed to the design and implementation of ADNI and/or provided data but did not participate in analysis or writing of this report. A complete listing of ADNI investigators can be found at: \url{http://adni.loni.usc.edu/wp-content/uploads/how_to_apply/ADNI_Acknowledgement_List.pdf}}
}
\end{abstract}
\maketitle

\section{Introduction}
\label{sec1}

To better understand and decode the information from genomic data, researchers often study the associations between the genetic variants and disease phenotypes. Recently, intermediate phenotypes are attracting more attention compared to the final disease diagnosis, since intermediate phenotypes potentially have stronger connections to the genetic variants and also provides comprehensive information for the final disease outcome. We consider data from the Alzheimer's Disease Neuroimaging Initiative (ADNI) study \citep{ADNI03} as an example.  It is a study for Alzheimer's disease with the intermediate multiple phenotypes being the structural magnetic resonance imaging (MRI) of the brain from the enrolled subjects, and many research efforts have been devoted in finding genetic variants that affect such phenotypes (\citet{ge2012increasing}, \citet{stein2010genome}, \citet{stein2010voxelwise}.  In this article, we will primarily focus on the problem of correlating the genetic variants with the imaging data.

One of the previous ADNI analyses applied linear regression using one genetic marker versus one phenotype (a single voxel of the brain MRI) in a massively univariate manner across all markers and all phenotypes \citep{stein2010voxelwise}. Such a method is feasible if the number of genetic variants and phenotypes is small; when the dimension of genotypes and phenotypes are both very large, the resulting test has limited power due to the issue of multiple comparisons. Two popular modelling frameworks that could potentially be applied to the motivating example are the multivariate kernel machine regression model (MV-KMR) \citep{maity2012multivariate} and the kernel distance covariance method (KDC) (\citet{Szekely07}, \citet{Szekely09}, and \citet{Gretton08}).

MV-KMR is a multivariate response framework based on kernel machine regression (KMR) \citep{maity2012multivariate}, where KMR is a semiparametric regression that models the covariates effect parametrically, and considers the genetic markers non-parametrically (\citet{liu2007semiparametric} and \citet{kwee2008powerful}). Specifically, the non-parametric effect of multiple markers is introduced by a kernel. The Gaussian RBF kernel is frequently used for quantitative measurements, while the IBS or polynomial kernel can be considered for the qualitative variables.  One advantage of this approach is that it is able to greatly simplify the specification of a non-parametric model for multiple markers effects \citep{liu2007semiparametric}. Since we focus our discussion on multivariate phenotypes in this work, we simply use KMR to denote the multivariate case, as the KMR model is feasible for both univariate and multiple phenotypes.

KDC is a term that we define as a class of methods for the tests of independence, and it includes distance covariance (DC) and Hilbert-Schmidt Independence Criterion (HSIC). DC was established by \citet{Szekely07} as providing a test of independence in high-dimensional settings that is consistent against all alternatives.  One advantage of distance covariance is the compact representation of the statistic which is the product of expectations of pairwise $L_2$ distance, which can be estimated empirically in a straightforward manner. On the other hand, \citet{Gretton08} formulated the two-variable independence test (HSIC) in Reproducing Kernel Hilbert Spaces (RKHS). Similar to DC, the HSIC statistic is able to find the test of dependence in multivariate spaces, and it is consistent when a characteristic kernel \citep{sriperumbudur2010hilbert} is used. \citet{sejdinovic2012equivalence} demonstrated that the above two statistics are the same when the distance-induced kernel in HSIC is chosen. The results in \citet{sejdinovic2012equivalence} showed that the HSIC test is more sensitive when the quantity is derived from other kernels, and the HSIC tests can be readily extended to more structured and non-Euclidean spaces.

In this work, we establish the equivalence between KMR and KDC. We begin our discussion by first reviewing two preliminaries in application to the linear model, followed by an algebraic representation of the multiple phenotypes version of KMR, and show that the KMR and KDC are equivalent under the condition of no covariate adjustment and when a Gower distance kernel\citep{gower1966some} or linear kernel is used for the phenotype spaces. Furthermore, we propose a new covariate-adjusted KDC test with the presence of the covariates, and show that KDC is equivalent to the KMR of \citet{maity2012multivariate}. Three major implications are found by the equivalence established in this work.  First, the equivalence shows that the principles of KMR can be applied to the interpretation of KDC.  Second, the new proposed covariate-adjusted KDC test shows an increase in power relative to the original KDC test  in our simulation studies.  Third, the KMR statistic is a member of the KDC family, in that the members are the quantities of different kernels. Our numerical results suggest that KDC may yield a more power result with kernels that are tailored to the particular application.

\section{Preliminaries}
\subsection{Distance properties of the sum of squares}
Before we introduce the equivalence between KMR and KDC, a review of the distance properties of the sums of squares \citep{gower1966some} is provided.  Suppose the multivariate sample has a set of $n$ points $\mathbf{Y}=(Y_1,...,Y_n)^t$ in $p-$dimensional space, where $Y_i$ has coordinates $(y_{i1},...,y_{ip})$.  Then the distance $d_{ij}$ between $Y_i$ and $Y_j$ is $d^2_{ij}=\sum^{p}_{r=1} (y_{ir}-y_{jr})^2.$ We denote $\mathbf{A}$ as a $n \times n$ symmetric sum of squares matrix of $\mathbf{Y}$ with eigenvalues $\lambda_1,...,\lambda_n$ and eigenvectors $\mathbf{c}_1,...,\mathbf{c}_n$. If $\mathbf{c}_1,...,\mathbf{c}_n$ are normalized, then $\mathbf{A} = \mathbf{c}_1\mathbf{c}_1'+\mathbf{c}_2\mathbf{c}_2'+...,\mathbf{c}_n\mathbf{c}_n'.$; where $a_{ii}=\sum^n_{r=1}c^2_{ir}$, and $a_{ij}=\sum^n_{r=1}c_{ir}c_{jr}$. Hence, the distance between $Y_i$ and $Y_j$ can be derived as follows:
\beqlb
\label{eq1}
d^2_{ij}&=&\sum^{p}_{r=1} (y_{ir}-y_{jr})^2 \nonumber \\
&=& \sum^{n}_{r=1}  (c_{ir}-c_{jr})^2 \nonumber \\
&=& a_{ii} + a_{jj} -2 a_{ij}.
\eeqlb
Now, $\mathbf{A}$ is centered by subtracting the means, then $a_{ii}=0$, and $a_{ij}=-\frac{1}{2}d^{2}_{ij}$ (denoted as the Gower distance \citep{gower1966some}), then the sum of squares matrix of $\mathbf{Y}$ is
\beqnn
 \mathbf{A}&=&\left (I - \frac{11'}{n} \right)\mathbf{Y}\mathbf{Y}'\left (I - \frac{11'}{n} \right)\\
            &=&-\frac{1}{2} d^{2}_{ij} (1_n 1'_n - I),
\eeqnn
where $1_n$ is a $n \times 1$ vector with elements 1, and $I$ is an identity matrix of size $n$.

\subsection{Linear model}
\label{s:MANOVA}
Linear models are often utilized to examine the effects between two variables. Suppose we observe $n$ subjects with index $i=1,...,n$, and the response $\mathbf{Y}=(Y_1,..,Y_n)^t$ in $p-$dimensional space and the predictor $\mathbf{Z}=(Z_1,...,Z_n)^t$ in $q-$dimensional space. To understand the relationships between $\mathbf{Y}$ and $\mathbf{Z}$, a typical way is to apply the multivariate linear model and in particular, MANOVA. Traditional multivariate analysis proceeds through partitioning of the total sum of squares $tr(\mathbf{Y}'\mathbf{Y})$, and the analysis can be done by the linear model $\mathbf{Y}=\mathbf{Z} \mathbf{\beta }+ \mathbf{\epsilon}$. For the test of the effects between $\mathbf{Y}$ and $\mathbf{Z}$, we formulate the hypothesis for testing $H_0: \mathbf{\beta} = 0$, and the least square estimates for $\beta$ is $\hat{\beta}=(\mathbf{Z'Z})^{-1}\mathbf{Z'Y}$. Therefore, the fitted values of $\mathbf{Y}$ is $\hat{\mathbf{Y}}=\mathbf{Z}\hat{\beta}=\mathbf{HY}$, where $\mathbf{H}=\mathbf{Z(Z'Z})^{-1}\mathbf{Z}$, and the sum of the residuals is $\mathbf{R}=\mathbf{Y}-\hat{\mathbf{Y}}$, and $tr(\mathbf{Y}'\mathbf{Y})=tr(\hat{\mathbf{Y}}'\hat{\mathbf{Y}})+tr(\mathbf{R}'\mathbf{R})$. Hence, an appropriate statistic to test the null hypothesis of the model having no effects is a pseudo $F$ statistic \citep{mcardle2001fitting}:
\beqlb
\label{eq2}
F=\frac{tr(\hat{\mathbf{Y}}'\hat{\mathbf{Y}})/(q-1)}{tr(\mathbf{R}'\mathbf{R})/(n-q)}=\frac{tr(\mathbf{H YY'H})/(q-1)}{tr(\mathbf{(I-H)YY'(I-H)})/(n-q)}.
\eeqlb
Furthermore, \citet{mcardle2001fitting} suggested that the above partitioning procedure can be done using the outer product matrix, i.e., $tr(\mathbf{YY}')$, since $tr(\mathbf{Y}'\mathbf{Y})=tr(\mathbf{YY}')$.
Therefore, we can replace $\mathbf{YY}'$ with any $n \times n$ distance matrix $\mathbf{D}$:
\beqlb
\label{eq3}
\frac{tr(\mathbf{HDH})/(q-1)}{tr(\mathbf{(I-H)D(I-H)})/(n-q)}.
\eeqlb

If $\mathbf{D}$ is a Gower distance matrix, then (\ref{eq3}) is the same as (\ref{eq2}); if $\mathbf{D}$ is some other distance (kernel) matrix, then the significance of (\ref{eq3}) can be tested using the permutation technique. The estimate in (\ref{eq3}) is flexible in capturing the nature of the response, and measures the effect between the predictors and the responses at the same time.

\section{Methods}
\citet{mcardle2001fitting} applied the outer product tool to extend the MANOVA to the general MANOVA. This inspires us to apply the same argument to the KMR model, and it turns out that KMR is equivalent to the KDC when a common kernel is chosen. To understand this link between KMR and KDC, we first consider the KMR model without covariates.  We then consider the situation when covariate effects are included, which leads to our proposal of a new covariate-adjusted KDC test with the presence of the covariates.

\subsection{Without covariates}

In order to test the dependence between two random vectors, i.e., the association between the phenotypes $\mathbf{Y}=(Y_1,...,Y_n)^t \in \mathbb{R}^p$ and the genotypes $\mathbf{Z}=(Z_1,...,Z_n)^t \in \mathbb{R}^q$, KDC (i.e., DC or HSIC) can be used for the purpose. Here we use the same algebraic formulation as the one in \citet{Gretton08}, and denote the kernel function $k_{ij}=k(Z_i,Z_j)$ as an element of row $i$ and column $j$ of the kernel matrix $K$ in $\mathbf{Z}$, and $l_{ij}=l(Y_i,Y_j)$ is the kernel function of $Y_i$ and $Y_j$ in the kernel matrix $L$ in $\mathbf{Y}$ space. Therefore, the KDC statistic for association between $\mathbf{Y}$ and $\mathbf{Z}$ is defined as
\beqlb
\label{eq12}
\mathrm{KDC}_n =\frac{1}{n^2} tr(KHLH) \propto tr(KHLH),
\eeqlb
where $tr(\mathbf{X})$ is the trace of $\mathbf{X}$ and $H=(I - {11'}/{n})$. If both $k$ and $l$ are $L_2$ distance kernels, then (\ref{eq12}) is the DC statistic \citep{Szekely07}; if other reproducing kernels are applied, (\ref{eq12}) is the HSIC statistic \citep{Gretton08}. In summary, the KDC statistic is used for testing the dependence between $\mathbf{Y}$ and $\mathbf{Z}$ without distributional assumptions.

Another powerful test for test of interactions is kernel machine regression, which we now briefly discuss and later relate to the KDC. The linear model in \citet{liu2007semiparametric} and \citet{kwee2008powerful} is given by
\beqlb
\label{eq5}
Y=\beta_0 + h(\mathbf{Z})+\mathbf{\epsilon},
\eeqlb
where $h(\cdot)$ is an unknown function to be estimated by the effects of the SNPs on the univariate response $Y$, and it is determined by a specified positive semi-definite kernel function $k(.,.)$. To test the effects $h(\cdot)$, \citet{liu2007semiparametric} proposed a hierarchical Gaussian process regression for the linear model (\ref{eq5}):
\beqnn
Y|h(\mathbf{Z}) \sim N\{\beta_0+h(\mathbf{Z}),\sigma^2\}, \ \ \ \ h(\cdot) \sim GP\{0, \tau K\}.
\eeqnn
Therefore, the null hypothesis is that phenotype $Y$ and the SNPs $\mathbf{Z}$ exhibit no association, and one can test $H_0:\tau =0$ since $h$ can be treated as the subject-specific random effects with mean 0 and covariance matrix $\tau K $. Thus, the corresponding variance component score test is proportional to:
\beqlb
\label{eq6}
Q &\propto&(Y-\bar{Y})'K(Y-\bar{Y}) \nonumber \\
&=&tr[(Y-\bar{Y})'K(Y-\bar{Y}) ] \nonumber \\
&=&tr[K(Y-\bar{Y})(Y-\bar{Y})' ] \nonumber \\
&=&tr[K (I - \frac{11'}{n}) YY'  (I - \frac{11'}{n}) ]\nonumber \\
&=&tr[K H YY' H ]
\eeqlb
By using the trace trick, (\ref{eq6}) can be extended into two directions: first, the previous work in \citet{liu2007semiparametric} and \citet{kwee2008powerful} focused on a single phenotype $Y$.  We can also replace $Y$ with a multivariate phenotype $\mathbf{Y}$, and it turns out that $tr[K H \mathbf{Y}\mathbf{Y}' H ]$ is equivalent to MV-KMR in \citet{maity2012multivariate} in the absence of covariates.  Second, a common kernel is used in $K$ for both KMR and KDC, and by replacing the outer product $YY'$ with any distance matrix $L$ in (\ref{eq6}) results in the equivalence of KMR and KDC in (\ref{eq12}).

\subsection*{With covariates}

In practice, we may want to know the relationship between the genotypes ($\mathbf{Z}$) and phenotypes ($\mathbf{Y}$) where the phenotypes are adjusted by the covariates ($\mathbf{X}$), and we observe $n$ samples from $\mathbf{X}\in \mathbb{R}^m,\mathbf{Y} \in \mathbb{R}^p,$ and $\mathbf{Z} \in \mathbb{R}^q$. Under this setting, the multivariate traits KMR model \citep{maity2012multivariate} is
\beqnn
\mathbf{Y}= \mathbf{X} \boldsymbol \beta  + h(\mathbf{Z}) +  \boldsymbol \epsilon,
\eeqnn
where $h(\cdot)$ is an non-parametric function which describes the effect of $\mathbf{Z}$. To test the effect of $h(\cdot)$, one can test $H_0: \tau_1=...\tau_p = 0$ under the following representation that is a multivariate extension of the hierarchical Gaussian process regression from the previous section:
\beqnn
\mathbf{Y}|( \boldsymbol \beta , h(\mathbf{Z})) \sim MVN\{\mathbf{X} \boldsymbol \beta + h(\mathbf{Z}),\Sigma \}, \ \ \ \ h(\cdot) \sim GP\{0, \boldsymbol \tau K\},
\eeqnn
and the corresponding score test of $H_0$ is proportional to
\beqlb
\label{eq9}
Q &\propto&(\mathbf{Y}- \mathbf{X}\hat{ \boldsymbol \beta})'K(\mathbf{Y}- \mathbf{X}\hat{\boldsymbol \beta}) \nonumber \\
&=&tr[(\mathbf{Y}- \mathbf{X}\hat{\boldsymbol \beta})'K(\mathbf{Y}-\mathbf{X}\hat{\boldsymbol \beta}) ] \nonumber \\
&=& tr[(\tilde{\mathbf{Y}}-\bar{\tilde{\mathbf{Y}}})'K(\tilde{\mathbf{Y}}-\bar{\tilde{\mathbf{Y}}}) ] \nonumber \\
&=&tr[K H\tilde{\mathbf{Y}}\tilde{\mathbf{Y}}' H ],
\eeqlb
where $\tilde{\mathbf{Y}}=\mathbf{Y} - \mathbf{X}\hat{\boldsymbol \beta}$, and $\bar{\tilde{\mathbf{Y}}}$ is the average of $\tilde{\mathbf{Y}}$ in (\ref{eq9}) with $H$ being a centering offset (normalized constant) $(I - {11'}/{n})$
Notice that $\hat{\boldsymbol \beta}$ is the MANOVA estimates in the linear model section.  Hence, (\ref{eq9}) is equivalent to the score test in KMR from \citet{maity2012multivariate}, and the outer product $\tilde{\mathbf{Y}}\tilde{\mathbf{Y}}'$ can be replaced with any distance measure $\tilde{L}$ so that (\ref{eq9}) becomes
\beqlb
\label{eq10}
tr[K H \tilde{L} H].
\eeqlb
The original KDC (i.e., DC in \citet{Szekely07} and HSIC in \citet{Gretton08}) was presented as a test of independence between $\mathbf{Y}$ and $\mathbf{Z}$.  Here, we extend it to the case when the covariates $\mathbf{X}$ are present. Specifically, by applying a common kernel on $K$ for both statistics, and if $\tilde{L}$ is composited by a Gowder distance kernel or a linear kernel, then KDC in (\ref{eq10}) is again equivalent to (\ref{eq9}) for the KMR.

For both cases when covariate effects are absent or present in (\ref{eq6}) and (\ref{eq9}), respectively, we ignore the covariance matrix structure of $\mathbf{Y}$ in order to establish the connection between KDC and KMR. Our idea for this step is similar to the work in \citet{pan2011relationship} that treats the covariance term as the fixed effects, while the covariance effects among $\mathbf{Y}$ are not skipped in our work. We demonstrate that the covariance effects of $\mathbf{Y}$ are able to be captured by choosing a suitable kernel matrix in the simulation studies.

\section{Simulation studies}
\subsection{Possible kernel choices}
There is a number of kernels for characterizing the similarity of individuals with respect to the variations of genotypes and phenotypes. We considered the following kernels for our numerical data analyses:

\begin{enumerate}
\item Identity-by-state (IBS) kernel: $k(Z_i,Z_j)= (2q)^{-1} \sum^{q}_{r=1} (2-| Z_{ir}-Z_{jr} |)$, where $q$ is the number of loci considered in the calculation.
\item Euclidean distance ($L_2$) kernel: $k(Z_i,Z_j)= \| Z_i-Z_j \|_q = \sqrt{\sum^{q}_{r=1} (Z_{ir}-Z_{jr})^2}$.
\item Gaussian RBF kernel: $k(Z_i,Z_j)= \exp\{ -\rho \| Z_i-Z_j \|^2_q \}$, where $\rho$ is the weight parameter.
\item Polynomial kernel: $k(Z_i,Z_j)=(\langle Z_i,Z_j \rangle +c)^{d}$, where $\langle Z_i,Z_j \rangle$ denotes the inner product of $Z_i$ and $Z_j$, and $c$ is a constant.\\
Notice that the polynomial kernel can be simplified into a linear kernel when $c$=0 and $d=1$, or into a quadratic kernel when $c=1$ and $d=2$.
\end{enumerate}

\subsection{Simulation studies}
The goal of the following simulations is to compare the performances of KMR and KDC in terms of the empirical size and powers under different kernel combinations.

\subsection{Simulation 1}
The first simulation examined the association between the effect of a single phenotype $Y$ and the multivariate $\mathbf{Z}$ adjusted by a single covariate $X$, and the design of the simulation was based on \citet{liu2007semiparametric}. The true linear model was:
\beqnn
Y = \beta_0+ \beta X  + h(Z_{1},...,Z_{q})+\epsilon,
\eeqnn
where $h(\mathbf{Z})=a h_1(\mathbf{Z})$, $h_1(\mathbf{Z})=2\cos(Z_1)-3Z^2_2+2 \exp(-Z_3)  Z_4 -1.6 \sin(Z_5)\cos(Z_3)+4 Z_1 Z_5$, and $X =3\cos(Z_{1}) + u$. The $Z_j's (j=1,...,5)$ were generated from uniform(0,1) while $u$ and $\epsilon$ were generated from independent $N(0,1)$.

To estimate for the coefficients of $X$, we first used the \textit{lm} function from \textsf{stat} package in R to solve $\hat{\beta}_0$ and $\hat{\beta}$, and then $\tilde{Y}=Y -\hat{\beta}_0 - \hat{\beta} X$. The empirical size and powers of the tests were evaluated by generating data under $a=0$ and $a=0.25,0.5,0.75,1.0$ at significance level of 0.05. The sample size was 60; the $p$-value of the statistic was computed based on $10^4$ permutations, and this experiment was repeated 1000 times. In the following, we used $K$ to represent the kernel matrix for the genotypes $\mathbf{Z}$, and $\tilde{L}$ to represent the kernel matrix for the adjusted phenotype $\tilde{Y}$.
Table \ref{t:s11} shows the results of empirical sizes and powers of KMR and KDC tests, where the linear and quadratic kernels were considered on $K$ of KMR, while the $L_2$ distance, linear and quadratic kernels were used in both $K$ and $\tilde{L}$ of KDC. As the results, the KMR and KDC were equivalent when  $\tilde{L}$ was a linear kernel in KDC, and the performances of the quadratic kernel resulted in less powers than its counterparts.

\begin{table}[!ht]
\centering
\caption{Empirical size and the powers of KMR and KDC with different choice of kernels: linear, quadratic, and $L_2$ distance. $K$ represents the kernel matrix for the genotypes $\mathbf{Z}$, and $\tilde{L}$ represents the kernel matrix for the adjusted phenotype $\tilde{Y}$}
\label{t:s11}
\begin{tabular}{l c c c c c c}
\Hline
& Size &&\multicolumn{4}{c}{Power} \\ [-1pt]
 \cline{2-2} \cline{4-7} \\ [-7pt]
	&	a=0	&	& a=0.25	&	a=0.5	&	a=0.75	&	a=1	\\[-1pt]
\hline
KMR ($K$=linear)&	0.044&	&	0.227	&	0.759	&	0.970	&	0.996	\\
KMR ($K$=quadratic)&	0.035&	&	0.198	&	0.706	&	0.962	&	0.995	\\
KDC ($\tilde{L}$,$K$=$L_2$)	&	0.049	&&	0.212	&	0.725	&	0.967	&	0.995	\\
KDC ($\tilde{L}$,$K$=linear)&	0.044&	&	0.227	&	0.759	&	0.970	&	0.996	\\
KDC ($\tilde{L}$=linear,$K$=quadratic)	&	0.035	&&	0.198	&	0.706	&	0.962	&	0.995	\\
KDC ($\tilde{L}$=quadratic,$K$=linear)	&	0.041	&&	0.173	&	0.585	&	0.877	&	0.933	\\
KDC ($\tilde{L}$,$K$=quadratic)	&	0.039	&&	0.146	&	0.519	&	0.822	&	0.908	\\
\hline
\end{tabular}
\end{table}

\subsection{Simulation 2}
For the second simulation, the design was based on \citet{maity2012multivariate}. For $k=1,...,p$, the data were generated through the model
\beqlb
\label{eq14}
Y_{k}=\mathbf{X} \boldsymbol \beta_{k} + h_k(\mathbf{Z}) + \epsilon_{k},
\eeqlb
where $\mathbf{X}=(X_{1},X_{2})^T$ were generated from bivariate normal $BVN((0.2,0.4)^T,I)$, and $\epsilon_{k}'s$ were generate $MVN(0,\Sigma_{true})$.
The $q$-SNP genotype data $\mathbf{Z} \equiv (Z_{1},...,Z_{q})$, with $q = 9$, were simulated as in the CATIE SNPs data described in  \citet{maity2012multivariate}. Two choices for the effects of $h_k (k=1,2,3)$ were considered.  The first is the sparse effect, i.e., $h_1= a (z_1+z_2+z_3+z_1z_4z_5-{z_6}/{3}-{z_7z_8}/{2}+(1-z_9)) $, $h_2=h_3=0$, where $a=0,0.1,0.2$.  The second is the common effect, i.e., $h^{\ast}_1=h_1 + a z_3$, and $h_2=h_3= a z_3$ with $a=0,0.1,0.2$. In addition, we also investigated the performances of KMR and KDC by varying the variance-covariance of $\Sigma_{true}$ using an independent structure ($\Sigma_1$) and a highly dependent structure ($\Sigma_2$):
\beqnn
\Sigma_1= \left( \begin{array}{ccc}
0.95 & 0 & 0 \\
0 & 0.86 & 0 \\
0 & 0 & 0.89 \end{array} \right)
\ \ \ \
\Sigma_2= \left( \begin{array}{ccc}
0.95 & 0.57 & 0.43 \\
0.57 & 0.86 & 0.24 \\
0.43 & 0.24 & 0.89 \end{array} \right)
\eeqnn
The empirical size ($a=0$) and powers ($a=0.1,0.2$) were examined at significance level of $0.05$. The sample size $n$ was 100, the dimension of genotypes $q$ was 9, and the dimension of phenotypes $p$ was 3. Furthermore, to adjust the covariate effects $\mathbf{X}$, we again used the \textit{lm} function from \textsf{stat} package in R to solve $\hat{\boldsymbol \beta}$, and then $\tilde{\mathbf{Y}}=\mathbf{Y}- \mathbf{X} \hat{\boldsymbol \beta} $, where $\mathbf{Y}=(Y_1,...,Y_p)$, and $\hat{\boldsymbol \beta}=\hat{\beta}_1,...,\hat{\beta}_p$. In this simulation, the linear, quadratic, IBS and $L_2$ distance kernels were used for the performance evaluation.

Table \ref{t:s21} shows the empirical size of KMR and KDC tests, and the values were all close to $\alpha =0.05$. This suggests that the tests are able to control the type I error. Table \ref{t:s22} displays the empirical powers of case 1 when the covariate effects are included (test of independence between $\mathbf{Z}$ and $\tilde{\mathbf{Y}}$), where the phenotypes $\tilde{\mathbf{Y}}$ are adjusted by the covariates; and case 2 when the covariate effects are excluded (test of independence between $\mathbf{Z}$ and $\mathbf{Y}$), where the phenotypes $\mathbf{Y}$ are the raw samples. The results showed that case 1 had greater powers than case 2, and it suggests that our proposed KDC test is able to incorporate the covariates and results in an increase in power; although the highly dependent structure of $\Sigma_2$ weakened the power estimates in Table \ref{t:s22}, the values remained very high. This implies that KDC test is able to identify the associations even when the phenotypes $\mathbf{Y}$ are correlated.

Overall, when the sparse effect and the independent covariance matrix $\Sigma_1$ were used, the best power performance was achieved when KMR and KDC were computed by a linear kernel on both $\tilde{L}$ and $K$ in Table \ref{t:s22}. This is expected because the linear kernel only identified the single connection between the $Y_1$ and $h_1$ without considering pairwise interactions. When the common effect and the dependent structure $\Sigma_2$ were considered, the KDC with the $L_2$ kernel on both $L$ and $K$ resulted in the largest empirical power. This suggests that the $L_2$ kernel is able to incorporate the dependent covariance structure, i.e., $\Sigma_2$, and also identify the interaction effects between $\mathbf{Y}$ and $\mathbf{Z}$ at the same time.

\begin{table}[!ht]
\centering
\caption{Empirical type I error rate (a=0) of KMR and KDC with the different choice of kernels: linear, quadratic, IBS, and $L_2$ distance. $K$ represents the kernel matrix for the genotypes $\mathbf{Z}$, and $\tilde{L}$ represents the kernel matrix for the adjusted phenotype $\tilde{Y}$}
\label{t:s21}
\begin{tabular}{l c c c c c c c}
\Hline
 &\multicolumn{2}{c}{Sparse effect }  &	 	&\multicolumn{2}{c}{Common effect}		\\	
\cline{2-3} \cline{5-6}\\[-7pt]
 	&	$\Sigma_1$	&	$\Sigma_2$	&	&	$\Sigma_1$	&	$\Sigma_2$	\\
 \hline
KMR ($K$=linear)	&	0.047	&	0.055	&	&	0.050	&	0.055	\\
KMR ($K$=quadratic)	&	0.047	&	0.056	&	&	0.047	&	0.059	\\
KMR ($K$=IBS)	&	0.043	&	0.060	&	&	0.050	&	0.059	\\
KDC ($\tilde{L}$,$K$=$L_2$)	&	0.036	&	0.059	&	&	0.045	&	0.061	\\
KDC ($\tilde{L}$,$K$=linear)	&	0.047	&	0.055	&	&	0.050	&	0.055	\\
KDC ($\tilde{L}$=linear,$K$=quadratic)	&	0.047	&	0.056	&	&	0.047	&	0.059	\\
KDC ($\tilde{L}$=linear,$K$=IBS)	&	0.043	&	0.060	&	&	0.050	&	0.059	\\
KDC ($\tilde{L}$=quadratic,$K$=linear)	&	0.043	&	0.057	&	&	0.052	&	0.052	\\
KDC ($\tilde{L}$,$K$=quadratic)	&	0.043	&	0.062	&	&	0.051	&	0.061	\\
KDC ($\tilde{L}$=quadratic,$K$=IBS)	&	0.034	&	0.066	&	&	0.052	&	0.061	\\
\hline
\end{tabular}
\end{table}

\begin{table}[!ht]
\centering
\caption{Powers (a=0.1, 0.2) of KMR and KDC of case 1: covariate effects included; and case 2: covariate effect excluded. The different choice of kernels are linear, quadratic, IBS, and $L_2$ distance for both cases. $K$ represents the kernel matrix for the genotypes $\mathbf{Z}$, and $\tilde{L}$ represents the kernel matrix for the adjusted phenotype $\tilde{Y}$. Note that $\tilde{.}$ represents the kernel matrices that are adjusted by the covariates.}
\label{t:s22}
\hspace*{-2cm}
\begin{tabular}{l c c c c c c c c c}
\Hline
 &\multicolumn{4}{c}{Sparse effect }  &	 	&\multicolumn{4}{c}{Common effect}		\\	
\cline{2-5} \cline{7-10}\\[-7pt]
& \multicolumn{2}{c}{$\Sigma_1$} & \multicolumn{2}{c}{$\Sigma_2$} 	&& \multicolumn{2}{c}{$\Sigma_1$}	 & \multicolumn{2}{c}{$\Sigma_2$}	\\
&	$a=$0.1	&	$a=$0.2	&	$a=$0.1	&	$a=$0.2	&&	$a=$0.1	&	$a=$0.2	&	$a=$0.1	&	$a=$0.2	\\
\hline
\multicolumn{10}{l}{Case 1: Covariate effects included}	\\ 		
KMR ($K$=linear)	&	0.410	&	0.984	&	0.327	&	0.973	&&	0.336	&	0.947	&	0.255	&	0.960	\\
KMR ($K$=quadratic)	&	0.405	&	0.980	&	0.322	&	0.974	&&	0.330	&	0.949	&	0.242	&	0.954	\\
KMR ($K$=IBS)	&	0.400	&	0.977	&	0.308	&	0.966	&&	0.330	&	0.946	&	0.237	&	0.946	\\
KDC ($\tilde{L}$,$K$=$L_2$)	&	0.364	&	0.970	&	0.355	&	0.982	&&	0.306	&	0.931	&	0.304	&	0.977	\\
KDC ($\tilde{L}$,$K$=linear)	&	0.410	&	0.984	&	0.327	&	0.973	&&	0.336	&	0.947	&	0.255	&	0.960	\\
KDC ($\tilde{L}$=linear,$K$=quadratic)	&	0.400	&	0.980	&	0.322	&	0.974	&&	0.330	&	0.949	&	0.242	&	0.954	\\
KDC ($\tilde{L}$=linear,$K$=IBS)	&	0.400	&	0.977	&	0.308	&	0.966	&&	0.330	&	0.946	&	0.237	&	0.946	\\
KDC ($\tilde{L}$=quadratic,$K$=linear)	&	0.252	&	0.922	&	0.171	&	0.756	&&	0.216	&	0.821	&	0.136	&	0.669	\\
KDC ($\tilde{L}$,$K$=quadratic)	&	0.253	&	0.926	&	0.179	&	0.748	&&	0.218	&	0.818	&	0.133	&	0.660	\\
KDC ($\tilde{L}$=quadratic,$K$=IBS)	&	0.258	&	0.908	&	0.177	&	0.757	&&	0.216	&	0.807	&	0.137	&	0.651	\\
\\[-7pt]
\multicolumn{10}{l}{Case 2: Covariate effect excluded}	\\ 			
KMR ($K$=linear)	&	0.336	&	0.947	&	0.242	&	0.930	&&	0.278	&	0.902	&	0.187	&	0.884	\\
KMR ($K$=quadratic)	&	0.326	&	0.954	&	0.237	&	0.927	&&	0.262	&	0.896	&	0.176	&	0.873	\\
KMR ($K$=IBS)	&	0.321	&	0.942	&	0.222	&	0.911	&&	0.261	&	0.886	&	0.172	&	0.863	\\
KDC ($L$,$K$=$L_2$)	&	0.314	&	0.930	&	0.266	&	0.964	&&	0.253	&	0.886	&	0.227	&	0.941	\\
KDC ($L$,$K$=linear)	&	0.336	&	0.947	&	0.242	&	0.930	&&	0.278	&	0.902	&	0.187	&	0.884	\\
KDC ($L$=linear,$K$=quadratic)	&	0.326	&	0.954	&	0.237	&	0.927	&&	0.262	&	0.896	&	0.176	&	0.873	\\
KDC ($L$=linear,$K$=IBS)	&	0.321	&	0.942	&	0.222	&	0.911	&&	0.261	&	0.886	&	0.172	&	0.863	\\
KDC ($L$=quadratic,$K$=linear)	&	0.201	&	0.876	&	0.144	&	0.779	&&	0.172	&	0.764	&	0.075	&	0.587	\\
KDC ($L$,$K$=quadratic)	&	0.206	&	0.884	&	0.142	&	0.786	&&	0.170	&	0.759	&	0.078	&	0.578	\\
KDC ($L$=quadratic,$K$=IBS)	&	0.213	&	0.875	&	0.138	&	0.774	&&	0.158	&	0.762	&	0.084	&	0.574	\\
\hline
\end{tabular}
\end{table}

\section{Experiments with the Alzheimer Disease Neuroimaging Initiative (ADNI) study}

In this section, we first briefly introduce the ADNI dataset, followed by our simulation design according to the ADNI data samples for examining the empirical size and powers, and we then apply KDC and KMR with different kernels to the real ADNI samples.

Data used in the preparation of this article were obtained from the Alzheimer Disease Neuroimaging Initiative (ADNI) database (adni.loni.usc.edu). One of the goals of the ADNI study is to perform genome-wide association tests on the entire genome, and identify the genetic variants that influence the voxel-level differences in brain MRI of the enrolled subjects. Several work have been published to investigate this goal, i.e., \citet{stein2010voxelwise}, \citet{stein2010genome}, \citet{shen2010whole}, \citet{vounou2010discovering}, and \citet{hibar2011voxelwise}. In the ADNI study, the multiple phenotypes are presented by brain structural magnetic resonance imaging (MRI) scans (31,662 brain voxels), each containing a value that represents the volumetric difference of a subject's brain MRI voxel from a healthy reference brain, and a tensor based morphometry is used to compute the 3D map of regional brain volume differences compared to an average template image based on healthy elderly subjects. The genotypes are encoded by 448,244 SNPs across the entire genome, and the demographic variables include gender and age of all the participants.

In this work, the phenotypes were summarized from 31,662 total voxels into 119 region-of-interests (ROIs), where the mapping of the ROIs was based on the GSK CIC Atlas \citep{Tziortzi11}. The average voxel volumetric differences within each region was used to represent each of the 119 ROIs. \citet{wenyu2013mutiple} used the DC test on the same 119 brain MRI regions, and discovered that the difference in brain volumes were highly associated with a common variant $rs11891634$ in the intron region of gene \emph{FLJ16124}, with a total of 141 SNPs within gene \emph{FLJ16124} that were identified by SNP-gene mapping from \citet{hibar2011voxelwise}. Furthermore, 741 of all subjects from the ADNI study passed the quality control filtering according to \citet{stein2010voxelwise} (206 normal older controls, 358 mild cognitive impairment (MCI) subjects, and 177 Alzheimer's Disease (AD) patients), which we retained for the simulation and real data analysis.

\subsection{Simulation based on ADNI}

For this simulation study, a linear model $\mathbf{Y} = \mathbf{X} \boldsymbol \beta  + h(\mathbf{Z})+ \boldsymbol \epsilon$ was used to find the associations between the multiple phenotypes and genotypes. Similar to simulation two, the correlated structure among the phenotypes was considered, and the design of phenotypes correlation was based on \citet{vounou2010discovering}: the authors suggest to use the frontal cortex regions according to the GSK CIC atlas, and estimate the pairwise correlations from those regions using MCI subjects from the ADNI data set. We followed the same procedure \citep{vounou2010discovering} and used eight ($p=8$) frontal cortex regions of 119 ROIs. (\ref{eq13}) shows the all-pairwise correlations ($r$) of eight frontal cortex regions that were based on the 358 MCI subjects: the left and right anterior dorsolateral prefrontal cortex (corresponds to row 1 and 2), posterior dorsolateral prefrontal cortex (row 3,4), anterior medial prefrontal cortex (row 5,6), and posterior medial prefrontal cortex (row 7,8).
\beqlb
\label{eq13}
r= \left( \begin{array}{ccccccccc}
1.00	&	0.95	&	0.97	&	0.87	&	0.53	&	0.97	&	-0.99	&	-0.87	\\
0.95	&	1.00	&	1.00	&	0.98	&	0.77	&	1.00	&	-0.90	&	-0.66	\\
0.97	&	1.00	&	1.00	&	0.96	&	0.72	&	1.00	&	-0.94	&	-0.72	\\
0.87	&	0.98	&	0.96	&	1.00	&	0.88	&	0.97	&	-0.81	&	-0.51	\\
0.53	&	0.77	&	0.72	&	0.88	&	1.00	&	0.73	&	-0.43	&	-0.04	\\
0.97	&	1.00	&	1.00	&	0.97	&	0.73	&	1.00	&	-0.93	&	-0.71	\\
-0.99	&	-0.90	&	-0.94	&	-0.81	&	-0.43	&	-0.93	&	1.00	&	0.92	\\
-0.87	&	-0.66	&	-0.72	&	-0.51	&	-0.04	&	-0.71	&	0.92	&	1.00	
\end{array} \right)
\eeqlb

We then estimated the eight ROIs' covariance matrix $\Sigma$ using 358 MCI participants. The 358 MCI samples were selected for the simulation design due to the relatively uniform MRI outputs in the MCI group \citep{vounou2010discovering}, and hence all $\epsilon'$s were generated from MVN$(0,\Sigma)$.
For the genotypes elements, all 141 SNPs on gene \emph{FLJ16124} were used, i.e. $\mathbf{Z} =(Z_{1},...,Z_{141})$. The effect of $h$ is defined as $h(Z_{1},...,Z_{141})=a \times h_1$, with only the first 5 SNPs, $(Z_1,...,Z_5)$ of 141 Z's were the causative SNPs, such that $h_1(Z_1,...,Z_5)=2\cos(Z_1)-3Z^2_2+2 \exp(-Z_3)  Z_4 -1.6 \sin(Z_5)\cos(Z_3)+4 Z_1 Z_5$. For the covariate effects, we considered gender and standardized age based on the same 358 MCI subjects, where gender ($X_1$) was generated from a Bernoulli distribution with $p=0.36$, and standardized age ($X_2$) was generated from a standard normal. A total of 100 samples were generated, and the empirical size ($a=0$) and powers ($a=0.05,0.1$) were computed based on $10^4$ permutation. This simulation was repeated 1000 times and the significance level was set as 0.05.

Table \ref{t:s31} shows the empirical size and powers results with the linear, quadratic, IBS and $L_2$ distance kernel of KMR and KDC, and the KDC test with the $L_2$ distance measure resulted in the highest power among all the kernels. We also implemented the KDC test with a Gaussian kernel for $\tilde{L}$ ($\rho \in 0.1, 0.5,1,5,10$), and the linear, quadratic, and IBS kernel for $K$ in Table \ref{t:s32}. The highest power estimate was observed when we used a Gaussian kernel for $\tilde{L}$ ($\rho$=0.1) and a linear kernel for $K$. The highest power in Table \ref{t:s32} is close to the power performance of KDC test with the $L_2$ distance kernel in Table \ref{t:s31}, which suggests that when the dimensions of phenotype and genotype are both very high, both the Gaussian RBF kernel (with optimal parameter $\rho$) and the $L_2$ distance kernel are able to describe the high dimensional interactions, and also result in powerful performances.
%

\begin{table}[!htb]
\centering
\caption{Empirical type I error rate (a=0) and powers (a=0.05, 0.1) of KMR and KDC with different choice of kernels: linear, quadratic, IBS and $L_2$ distance. $K$ represents the kernel matrix on the genotypes $\mathbf{Z}$, and $\tilde{L}$ represents the kernel matrix on the adjusted phenotype $\tilde{Y}$}
\label{t:s31}
\begin{tabular}{l c c c c c c c }
\Hline
& \multicolumn{1}{c}{Size} & & \multicolumn{2}{c}{Power}\\
\cline{2-2} \cline{4-5}\\[-7pt]
 &	$a=0.0$	&&	$a=0.05$	&	$a=0.1$	\\
\hline
KMR ($K$=linear)	&	0.051	&&	0.270	&	0.928	\\
KMR ($K$=quadratic)	&	0.052	&&	0.243	&	0.887	\\
KMR ($K$=IBS)	&	0.058	&&	0.232	&	0.875	\\
KDC ($\tilde{L}$,$K$=$L_2$)	&	0.046	&& 0.378	&	0.974	\\
KDC ($\tilde{L}$,$K$=linear)	&	0.051	&&	0.270	&	0.928	\\
KDC ($\tilde{L}$=linear,$K$=quadratic)	&	0.052	&&	0.243	&	0.887	\\
KDC ($\tilde{L}$=linear,$K$=IBS)	&	0.058	&&	0.232	&	0.875	\\
KDC ($\tilde{L}$=quadratic,$K$=linear)	&	0.056	&&	0.067	&	0.145	\\
KDC ($\tilde{L}$,$K$=quadratic)	&	0.054	&&	0.058	&	0.121	\\
KDC ($\tilde{L}$=quadratic,$K$=IBS)	&	0.055	&&	0.068	&	0.165	\\
\hline
\end{tabular}
\end{table}

\begin{sidewaystable}[!ht]
\centering
\caption{Empirical type I error rate (a=0) and powers (a=0.05, 0.1) of the KDC test with a Gaussian RBF kernel for $\tilde{L}$ and the linear, quadratic, or IBS kernel for $K$}
\label{t:s32}
\hspace*{-3.2cm}
\begin{tabular}{l c c c c c c c c c c c c c c c c}
\Hline
&   \multicolumn{3}{c}{Size} &&   \multicolumn{7}{c}{Power}	\\[1pt]
\cline{2-4} \cline{6-12}\\ [-7pt]
&\multicolumn{3}{c}{$a=0.0$} && \multicolumn{3}{c}{$a=0.05$}  &&\multicolumn{3}{c}{$a=0.1$} \\
\cline{2-4} \cline{6-8} \cline{10-12}\\ [-7pt]
&	Linear	&	Quadratic	&	IBS &&	Linear	&	Quadratic	&	IBS	&&	Linear	&	Quadratic	&	IBS	\\
\hline
0.1	&	0.050	&	0.054	&	0.047	&	&	0.423	&	0.374	&	0.379	&	&	0.976	&	0.949	&	0.948	\\
0.5	&	0.046	&	0.046	&	0.047	&	&	0.469	&	0.429	&	0.422	&	&	0.899	&	0.854	&	0.893	\\
1	&	0.046	&	0.045	&	0.042	&	&	0.481	&	0.428	&	0.434	&	&	0.781	&	0.732	&	0.803	\\
5	&	0.039	&	0.036	&	0.040	&	&	0.315	&	0.288	&	0.337	&	&	0.389	&	0.337	&	0.437	\\
10	&	0.049	&	0.044	&	0.037	&	&	0.219	&	0.211	&	0.256	&	&	0.261	&	0.229	&	0.308	\\
\hline
\end{tabular}
\end{sidewaystable}

\subsection{Real data analysis}

The KMR and KDC tests were conducted using real ADNI samples to find the associations between the genetic variants and the multivariate brain MRI voxel with the demographic effects of gender and age. In contrast to the previous simulation setup, we utilized all 741 subjects of the ADNI study, 141 SNPs within gene \emph{FLJ16124}, 119 ROIs, and two covariates, i.e., gender and age. Table \ref{t:r1} displays the $p$-value of KMR and KDC with different choice of kernels (linear, quadratic, IBS, and $L_2$ distance kernel), where $p$-values were based on $10^4$ permutations. KMR ($K$=quadratic) and KDC ($\tilde{L}$=linear,$K$=quadratic) identified the smallest $p$-value, which suggests the quadratic kernel on the 141 SNPs has the most powerful performance in the KDC family among all the different choices of kernels. This suggests that there exists strong pairwise interaction effects among the SNPs in \emph{FLJ16124} that associate with brain region volumes, adjusting for age and gender.

\begin{table}[!htb]
\centering
\caption{Real data results: p-values of KMR and KDC with different choice of kernels}
\label{t:r1}
\begin{tabular}{l c}
\Hline
Test & p-values\\
\hline
KMR ($K$=linear)	&	0.063	\\
KMR ($K$=quadratic)	&	0.029	\\
KMR ($K$=IBS)	&	0.267	\\
KDC ($\tilde{L}$,$K$=$L_2$)	&	0.071	\\
KDC ($\tilde{L}$,$K$=linear)	&	0.063	\\
KDC ($\tilde{L}$=linear,$K$=quadratic)	&	0.029	\\
KDC ($\tilde{L}$=linear,$K$=IBS)	&	0.267	\\
KDC ($\tilde{L}$=quadratic,$K$=linear)	&	0.073	\\
KDC ($\tilde{L}$,$K$=quadratic)		&	0.035	\\
KDC ($\tilde{L}$=quadratic,$K$=IBS) &	0.270\\
\hline
\end{tabular}
\end{table}

\section{Conclusion and discussion}
In this work, we provided the algebraic formulation that the score of KMR is equivalent to the KDC statistic. The advantage of such equivalence allows the use of KMR interpretation to explain the KDC test. For instance, \citet{liu2007semiparametric} and \citet{maity2012multivariate} provided the REML score test of KMR, and it suggests that the null distribution of the KMR test is able to fit on the KDC test for the hypothesis testing, which greatly reduces computational costs compared to the permutation approach. Exploring the parametric distribution of the KDC statistic deserves further investigation, but is beyond the scope of this article.

In addition, the KMR tests can be treated as the members of a larger family of KDC, and more powerful test can be designed by looking at the optimal kernel among the family members. Although there is no single kernel that was concluded as the best from our simulations, the linear kernel for KMR or KDC achieved better performances than other kernels when considering the single phenotype; when the multiple phenotypes with multiple correlations or dependent covariance were presented, the Gaussian RBF or the $L_2$ kernel achieved better performances than other kernels. This result can be extended into designing a strategy to select for the the optimal kernel from the members of the large KDC family.

Finally, several work have utilized the KDC/KMR family members in the applications including genetic pathway analysis using KMR \citep{liu2007semiparametric}, voxel-wise genome-wide association studies using least square KMR \citep{ge2012increasing}, Neuroimaging genome-wide association using DC \citep{wenyu2013mutiple}, and multiple change point analysis using DC by \citet{matteson2013nonparametric}. Two recent work have presented and discussed the equivalence between the family members of our work, such as the relationships between Genomic Distance-Based Regression (GDBR) and KMR from \citet{pan2011relationship} and the equivalence between DC and HSIC from \citet{sejdinovic2012equivalence}. Therefore, our establishment of the equivalence between KMR and KDC in this work is an important unification of all the above applications. 

\section*{Acknowledgements}
Data collection and sharing for this project was funded by the Alzheimer's Disease Neuroimaging Initiative
(ADNI) (National Institutes of Health Grant U01 AG024904) and DOD ADNI (Department of Defense award
number W81XWH-12-2-0012). ADNI is funded by the National Institute on Aging, the National Institute of
Biomedical Imaging and Bioengineering, and through generous contributions from the following: Alzheimer's
Association; Alzheimer's Drug Discovery Foundation; BioClinica, Inc.; Biogen Idec Inc.; Bristol-Myers
Squibb Company; Eisai Inc.; Elan Pharmaceuticals, Inc.; Eli Lilly and Company; F. Hoffmann-La Roche Ltd
and its affiliated company Genentech, Inc.; GE Healthcare; Innogenetics, N.V.; IXICO Ltd.; Janssen
Alzheimer Immunotherapy Research \& Development, LLC.; Johnson \& Johnson Pharmaceutical Research
\& Development LLC.; Medpace, Inc.; Merck \& Co., Inc.; Meso Scale Diagnostics, LLC.; NeuroRx
Research; Novartis Pharmaceuticals Corporation; Pfizer Inc.; Piramal Imaging; Servier; Synarc Inc.; and
Takeda Pharmaceutical Company. The Canadian Institutes of Health Research is providing funds to support ADNI clinical sites in Canada.
Private sector contributions are facilitated by the Foundation for the
National Institutes of Health (www.fnih.org). The grantee organization is the Northern California Institute for
Research and Education, and the study is coordinated by the Alzheimer's Disease Cooperative Study at the
University of California, San Diego. ADNI data are disseminated by the Laboratory for Neuro Imaging at the
University of Southern California.
\backmatter




\bibliographystyle{biom} \bibliography{refs}

\label{lastpage}

\end{document}